\pdfoutput=1
\documentclass[11pt]{article}

\usepackage[final]{coling}

\usepackage{times}
\usepackage{latexsym}

\usepackage[T1]{fontenc}
\usepackage{tabularx} % For tables that adjust width
\usepackage{graphicx}
\usepackage{multirow}
\usepackage[utf8]{inputenc}

\usepackage{microtype}

\usepackage{inconsolata}

\usepackage{graphicx}
\usepackage{amsmath}

\title{LuxVeri at GenAI Detection Task 3: Cross-Domain Detection of AI-Generated Text Using Inverse Perplexity-Weighted Ensemble of Fine-Tuned Transformer Models}

\author{
 \textbf{Md Kamrujjaman Mobin\textsuperscript{1}},
 \textbf{Md Saiful Islam\textsuperscript{1, 2}} 
\\
 \textsuperscript{1}Computer Science and Engineering, Shahjalal University of Science and Technology,\\ Sylhet, Bangladesh\\
 \textsuperscript{2}Computing Science,  University of Alberta, Edmonton, Alberta, Canada\\
}

\begin{document}
\maketitle
\begin{abstract}
This paper presents our approach for Task 3 of the GenAI content detection workshop at COLING-2025, focusing on Cross-Domain Machine-Generated Text (MGT) Detection. We propose an ensemble of fine-tuned transformer models, enhanced by inverse perplexity weighting, to improve classification accuracy across diverse text domains. For Subtask A (Non-Adversarial MGT Detection), we combined a fine-tuned RoBERTa-base model with an OpenAI detector-integrated RoBERTa-base model, achieving an aggregate TPR score of 0.826, ranking 10th out of 23 detectors. In Subtask B (Adversarial MGT Detection), our fine-tuned RoBERTa-base model achieved a TPR score of 0.801, securing 8th out of 22 detectors. Our results demonstrate the effectiveness of inverse perplexity-based weighting for enhancing generalization and performance in both non-adversarial and adversarial MGT detection, highlighting the potential for transformer models in cross-domain AI-generated content detection.
\end{abstract}

\section{Introduction}

The proliferation of advanced language models such as GPT \citep{radford2019language} and RoBERTa \citep{liu2019roberta}, machine-generated content has become prevalent across social media, journalism, and academia, raising concerns about authenticity and misinformation. Detecting AI-generated text is especially challenging across diverse domains, where variations in language and style can hinder detection efforts.

In Task 3 of the COLING 2025 Workshop on Detecting AI-Generated Content \citep{dugan-etal-task3-overview}, we tackle cross-domain Machine-Generated Text (MGT) detection using an ensemble approach that combines fine-tuned RoBERTa-base models \citep{liu2019roberta} and OpenAI detection tools \citep{solaiman2019release}. Our method leverages inverse perplexity weighting to enhance the contributions of high-confidence models, yielding a robust detection system.

Our approach achieved an aggregate score of 0.826 in Non-Adversarial Cross-Domain MGT detection (Subtask A), ranking 10th, and 0.801 in Adversarial Cross-Domain MGT detection (Subtask B), ranking 8th. This paper outlines our ensemble-based methodology, dataset considerations, and insights for effective cross-domain AI-generated text detection.

\begin{table*}[h]
    \centering
    \small
    \renewcommand{\arraystretch}{1.1} % Adjust the row spacing
    \setlength{\tabcolsep}{1.5pt} % Adjust column separation for clarity
    \begin{tabular}{p{2cm}|*{8}{>{\centering\arraybackslash}p{1.4cm}}|>{\centering\arraybackslash}p{1.4cm}}
        \hline
        \hline
        \multirow{2}{*}{\textbf{Model}} & \multicolumn{8}{c|}{\textbf{Domain}} & \textbf{Total} \\
        \cline{2-10}
        & \textbf{Abstracts} & \textbf{Books} & \textbf{News} & \textbf{Poetry} & \textbf{Recipes} & \textbf{Reddit} & \textbf{Reviews} & \textbf{Wiki} & \\
        \hline \hline
        \textbf{Human} & 2119 & 2137 & 2136 & 2125 & 2126 & 2135 & 1132 & 2135 & 17109 \\
        \hline
        \textbf{ChatGPT} & 4238 & 4274 & 4272 & 4250 & 4253 & 4270 & 2263 & 4270 & 34090 \\
        \hline
        \textbf{Cohere} & 4238 & 4274 & 4272 & 4250 & 4253 & 4270 & 2263 & 4270 & 34090 \\
        \hline
        \textbf{Cohere-Chat} & 4238 & 4274 & 4272 & 4250 & 4253 & 4270 & 2263 & 4270 & 34090 \\
        \hline
        \textbf{GPT-2} & 8477 & 8549 & 8544 & 8501 & 8506 & 8540 & 4526 & 8540 & 68183 \\
        \hline
        \textbf{GPT-3} & 4238 & 4274 & 4272 & 4250 & 4253 & 4270 & 2263 & 4270 & 34090 \\
        \hline
        \textbf{GPT-4} & 4238 & 4274 & 4272 & 4250 & 4253 & 4270 & 2263 & 4270 & 34090 \\
        \hline
        \textbf{Llama-Chat} & 8477 & 8549 & 8544 & 8501 & 8506 & 8540 & 4526 & 8540 & 68183 \\
        \hline
        \textbf{Mistral} & 8477 & 8549 & 8544 & 8501 & 8506 & 8540 & 4526 & 8540 & 68183 \\
        \hline
        \textbf{Mistral-Chat} & 8477 & 8549 & 8544 & 8501 & 8506 & 8540 & 4526 & 8540 & 68183 \\
        \hline
        \textbf{MPT} & 8477 & 8549 & 8544 & 8501 & 8506 & 8540 & 4526 & 8540 & 68183 \\
        \hline
        \textbf{MPT-Chat} & 8477 & 8549 & 8544 & 8501 & 8506 & 8540 & 4526 & 8540 & 68183 \\
        \hline
        \hline
        \textbf{Total} & 62516 & 63577 & 63538 & 63380 & 63408 & 63606 & 37338 & 63606 & 518469 \\
        \hline
    \end{tabular}
    \caption{Data distribution for various models across different domains, with total data per model and summed values for each domain. The values represent domain-specific sample sizes for each model. We used only 10\% of the RAID \cite{dugan-etal-2024-raid} dataset for fine-tuning our models.}
    \label{tab:data_distribution_reduced}
\end{table*}

\section{Background}
\subsection{Dataset}
The RAID dataset \cite{dugan-etal-2024-raid}, provided for the competition, is designed for evaluating machine-generated text detectors. It contains over 10 million documents across 11 language models, 11 genres, 4 decoding strategies, and 12 adversarial attacks, including both human-written and machine-generated content from 8 different domains like books, news, poetry, and recipes. For training and validation, we used the RAID-train subset (802 million words, 11.8GB) and RAID-test subset (81 million words, 1.22GB). We also utilized the RAID-extra subset, which includes languages like Czech and German (275 million words, 3.71GB). This dataset provides a comprehensive resource for AI-generated text detection.

For the fine-tuning of our model, we reduced the dataset by using about 10\% of the publicly available data. This reduction was carried out in a balanced manner across all genres, decoding strategies, attacks, and domains to ensure that each subset was proportionally represented. Specifically, we reduced the data across the following domains: abstracts, books, news, poetry, recipes, reddit, reviews, and wiki. The distribution of this reduced data across models is shown in Table \ref{tab:data_distribution_reduced}, with domain-specific sample sizes for each model. For example, the number of samples for "ChatGPT" in the "Books" domain is 4274, while for "Human" in the "Reviews" domain, it is 1132. This balanced reduction ensures the data used for training is representative across models and domains, enabling efficient and effective fine-tuning.

\subsection{Related Work}

The detection of machine-generated text has gained attention with the rise of large language models (LLMs) like GPT \cite{radford2019language} and BERT \cite{devlin2019bert}. Fine-tuned Transformer models have succeeded in binary classification tasks, but challenges remain in cross-domain and multilingual contexts due to data biases \cite{liu2019roberta, solaiman2019release}. Ensemble methods combining models like BERT, RoBERTa, GPT variants, and perplexity-based weighting have been explored to improve domain robustness \cite{schick2020exploiting, clark2019electra}.

Recent work in cross-domain detection shows that RoBERTa-based detectors for GPT-2 generated technical text can be transferred with few labeled examples, such as from physics to biomedicine \cite{rodriguez-etal-2022-cross}. Paragraph-level detection is also being explored to address document tampering in mixed-domain texts.

For multilingual detection, models like XLM-RoBERTa \cite{DBLP:journals/corr/abs-1911-02116} and RemBERT \cite{DBLP:conf/iclr/ChungFTJR21} improve cross-lingual detection, though challenges remain for low-resource languages \cite{hu2020xtreme}. Recent SemEval tasks \cite{fetahu-etal-2023-semeval, wang2024semeval2024task8multidomain} have refined these approaches with task-specific fine-tuning. Our work builds on these methods by using inverse perplexity-weighted ensembles to enhance detection across domains and languages.

\begin{figure}[h]
    \centering
    \includegraphics[width=1\columnwidth]{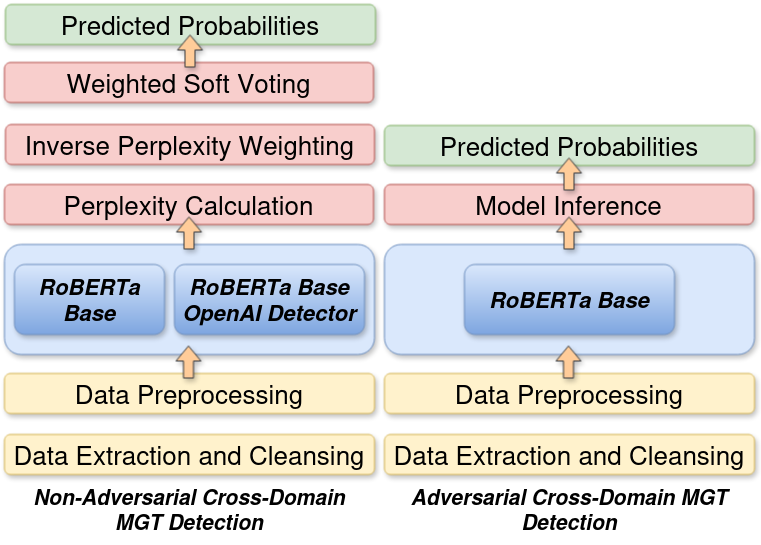}
    \caption{Overall Framework of our Proposed System for both Non-Adversarial and Adversarial Cross-Domain MGT Detection.}
    \label{fig:sys_overview}
\end{figure}

\section{System Overview}
We developed an ensemble approach for AI-generated text detection across multiple domains, using Transformer models with inverse perplexity-based weighted voting for improved accuracy. The system overview is shown in Figure \ref{fig:sys_overview}.

\subsection{Ensemble Model Selection and Justification}
For the ensemble model, we selected two Transformer-based models tailored for Non-Adversarial and Adversarial cross-domain text detection, leveraging their strengths in capturing linguistic, syntactic, and semantic patterns essential for AI-generated content detection. For non-adversarial tasks, we employed \textbf{RoBERTa-base}, recognized for its strong performance in natural language understanding and robust generalization across text domains \cite{liu2019roberta}, alongside the \textbf{RoBERTa-base OpenAI Detector}, fine-tuned specifically for distinguishing AI-generated text from human-authored content \cite{solaiman2019release}. For adversarial scenarios, \textbf{RoBERTa-base} was selected again for its ability to capture subtle linguistic patterns and semantic inconsistencies, making it highly effective in challenging detection tasks \cite{liu2019roberta}. This strategic selection ensures a balanced approach to handling diverse and adversarial text detection challenges.

% For the ensemble, we selected two Transformer-based models for Non-Adversarial and Adversarial cross-domain text detection, leveraging their ability to capture linguistic, syntactic, and semantic patterns for AI-generated content detection.

% \begin{itemize} \item \textbf{Non-Adversarial Models}: \begin{itemize} \item \textbf{RoBERTa-base}: Known for strong performance in natural language understanding and robust generalization across text domains \cite{liu2019roberta}. \item \textbf{RoBERTa-base OpenAI Detector}: Fine-tuned for AI-generated text detection, enhancing its ability to distinguish between human and AI content \cite{solaiman2019release}. \end{itemize}

% \item \textbf{Adversarial Model}:  
% \begin{itemize}  
%     \item \textbf{RoBERTa-base}: Excels in adversarial scenarios by capturing linguistic patterns and semantic inconsistencies \cite{liu2019roberta}.
% \end{itemize}  

% \end{itemize}

\subsection{Data Pre-processing}  
For text classification, the data was preprocessed using model-specific tokenizers, incorporating truncation and padding as required. To enhance memory efficiency and training performance, texts were sorted by word count, reducing unnecessary padding. A fixed random seed was maintained to ensure reproducibility.

\subsection{Training Procedure}  
The models were fine-tuned using the Hugging Face Transformers library \footnote{Hugging Face Transformers: \url{https://huggingface.co/transformers/}} for English and multilingual text classification. Tokenization was performed with `AutoTokenizer`, and the architectures were adapted for classification tasks with appropriate label mappings.

Training was conducted for 3 epochs with a learning rate of $2 \times 10^{-5}$, batch sizes of 4 for training and 16 for validation, and weight decay of 0.01. Early stopping was applied with a patience of 5 evaluations and a 0.001 improvement threshold. Evaluation checkpoints were saved after each epoch, and the best-performing model was used for testing.

This procedure ensured robust generalization across subtasks. Further training details are provided in Table \ref{table:training_config}.

\begin{table}[h]
\centering
\small
\renewcommand{\arraystretch}{1.3} % Adjust the row spacing
\setlength{\tabcolsep}{3pt} % Adjust column separation
\begin{tabular}{>{\raggedright\arraybackslash}p{3.5cm}|>{\raggedright\arraybackslash}p{3.5cm}} % Adjust column width and centering
\hline \hline
\textbf{Hyperparameter} & \textbf{Value} \\ \hline 
Number of Epochs        & $2\sim3$ \\ 
Learning Rate           & $1 \times 10^{-5} \sim 2 \times 10^{-5}$ \\ 
Training Batch Size     & 4 \\ 
Validation Batch Size   & 16 \\ 
Early Stopping Patience & 5 validation steps \\ 
Early Stopping Threshold& 0.001 \\ 
Weight Decay            & 0.01 \\ 
Optimizer               & AdamW \\ 
Loss Function           & Binary Cross-Entropy \\ 
Evaluation Strategy     & Every ¼ epoch \\ 
Checkpointing Strategy  & Validation loss \\ 
\end{tabular}
\caption{Training Configuration}
\label{table:training_config}
\end{table}

\begin{table*}[h]
    \centering
    \small
    \renewcommand{\arraystretch}{1.4} % Adjust the row spacing
    \setlength{\tabcolsep}{1.3pt} % Adjust column separation for clarity
    \begin{tabular}{>{\centering\arraybackslash}p{1.9cm}|>{\centering\arraybackslash}p{1cm}|>{\centering\arraybackslash}p{1cm}|>{\centering\arraybackslash}p{1cm}|>{\centering\arraybackslash}p{1cm}|>{\centering\arraybackslash}p{1cm}|>{\centering\arraybackslash}p{1.1cm}|>{\centering\arraybackslash}p{1cm}|>{\centering\arraybackslash}p{1.1cm}|>{\centering\arraybackslash}p{1.1cm}|>{\centering\arraybackslash}p{0.8cm}|>{\centering\arraybackslash}p{0.9cm}|>{\centering\arraybackslash}p{0.9cm}|>{\centering\arraybackslash}p{0.7cm}}
        \hline
        \multicolumn{14}{c}{\textbf{Non-Adversarial Results}} \\
        \hline
        \textbf{Detector} & \textbf{Chat GPT} & \textbf{GPT-4} & \textbf{GPT-3} & \textbf{GPT-2} & \textbf{Mistral} & \textbf{Mistral-Chat} & \textbf{Cohere} & \textbf{Cohere-Chat} & \textbf{Llama-Chat} & \textbf{MPT} & \textbf{MPT-Chat} & \textbf{AGG TPR} & \textbf{Rank} \\
        \hline \hline
        FT RoBERTa + RoBERTa OpenAI & 0.960 & 0.861 & 0.895 & \textbf{0.753} & \textbf{0.734} & 0.936 & 0.546 & 0.748 & 0.891 & \textbf{0.804} & 0.901 & \textbf{0.826} & 10/23 \\
        \hline
        FT RoBERTa + RoBERTa OpenAI + BERT & 0.983 & \textbf{0.934} & 0.755 & 0.730 & 0.709 & \textbf{0.960} & 0.510 & 0.747 & 0.943 & 0.772 & \textbf{0.932} & 0.825 & 11/23 \\
        \hline
        FT RoBERTa & 0.943 & 0.836 & 0.902 & 0.739 & 0.719 & 0.916 & 0.542 & 0.737 & 0.870 & 0.800 & 0.891 & 0.813 & 12/23 \\
        \hline
        Binoculars & \textbf{0.997} & 0.907 & \textbf{0.989} & 0.678 & 0.610 & 0.914 & \textbf{0.935} & \textbf{0.943} & \textbf{0.973} & 0.447 & 0.707 & 0.790 & - \\
        \hline
        \multicolumn{14}{c}{\textbf{Adversarial Results}} \\
        % \hline
        % \textbf{Generator Model} & \textbf{ChatGPT} & \textbf{GPT-4} & \textbf{GPT-3} & \textbf{GPT-2} & \textbf{Mistral} & \textbf{Mistral-Chat} & \textbf{Cohere} & \textbf{Cohere-Chat} & \textbf{Llama-Chat} & \textbf{MPT} & \textbf{MPT-Chat} & \textbf{AGG} & \textbf{Rank} \\
        \hline \hline
        FT RoBERTa & 0.911 & 0.808 & \textbf{0.873} & \textbf{0.730} & \textbf{0.720} & 0.887 & \textbf{0.567} & \textbf{0.740} & 0.855 & \textbf{0.806} & \textbf{0.861} & \textbf{0.801} & 8/22 \\
        \hline
        FT RoBERTa + RoBERTa OpenAI & 0.876 & 0.777 & 0.813 & 0.690 & 0.681 & 0.851 & 0.518 & 0.696 & 0.823 & 0.757 & 0.817 & 0.760 & 10/22 \\
        \hline
        FT RoBERTa + RoBERTa OpenAI + BERT & 0.896 & 0.843 & 0.675 & 0.663 & 0.651 & 0.874 & 0.457 & 0.670 & 0.857 & 0.711 & 0.841 & 0.749 & 11/22 \\
        \hline
        SuperAnnotate AI Detector & \textbf{0.963} & \textbf{0.913} & 0.720 & 0.411 & 0.342 & \textbf{0.897} & 0.445 & 0.685 & \textbf{0.918} & 0.314 & 0.767 & 0.649 & - \\
        \hline
    \end{tabular}
    \caption{Cross-domain MGT detection performance under non-adversarial and adversarial conditions. The table shows detector performance across various generator models, with aggregate True Positive Rate (AGG TPR) and rankings. "FT" denotes fine-tuned models, and base models are used for training and evaluation.}
    \label{tab:combined_results}
\end{table*}

\subsection{Ensemble Voting Strategy}

Our ensemble employs a weighted soft-voting strategy, combining predictions from all fine-tuned models for each subtask. The weights are determined based on inverse perplexity, with lower perplexity values reflecting higher confidence.

\subsubsection{Perplexity Calculation}
For each model, we compute the perplexity based on its predictions. The perplexity \( P \) is computed using the Negative Log Likelihood formula:

\[
P = \exp \left( - \frac{1}{N} \sum_{i=1}^{N} \log(p(y_i \mid x_i)) \right)
\]

where \( p(y_i \mid x_i) \) is the predicted probability for the true label \( y_i \), and \( N \) is the number of test samples. Lower perplexity values indicate higher confidence.

To compute perplexity, we use each model’s logits, apply softmax to obtain probabilities, and then calculate perplexity based on the true labels and these probabilities.

\subsubsection{Perplexity-Based Weighting Adjustment}

To calculate model weights, each model's perplexity is adjusted by subtracting 1, creating an effective weighting scale. The weight \( w_i \) for model \( i \) is then computed as the inverse of this adjusted perplexity and normalized across models, giving higher confidence models greater influence.

\[
w_i = \frac{1 / (P_i - 1)}{\sum_{j=1}^{M} (1 / (P_j - 1))}
\]

where \( M \) represents the total number of models, and \( P_i \) is the original perplexity of model \( i \).

\subsubsection{Weighted Soft-Voting}
Each model’s predicted probabilities are scaled by its weight and summed to form the final ensemble prediction. This weighted voting prioritizes models with higher confidence (lower perplexity), giving them greater influence on the final decision. The ensemble’s final prediction for each class \( c \) is:

\[
p_{\text{ensemble}}(c) = \sum_{i=1}^{M} w_i \cdot p_i(c)
\]

where \( p_i(c) \) is the predicted probability for class \( c \) by model \( i \), and \( w_i \) is its weight.

This method enhances ensemble accuracy by prioritizing predictions from more confident models, improving overall performance.

\section{Results}

Table \ref{tab:combined_results} shows cross-domain MGT detection performance for non-adversarial and adversarial testing, with detectors ranked based on aggregate True Positive Rate (TPR).

\subsection{Performance}

In the non-adversarial setting, the fine-tuned RoBERTa + RoBERTa OpenAI model which was fine-tuned on RAID dataset \cite{dugan-etal-2024-raid} achieved the highest performance, with an aggregate (AGG) score of 0.826, ranking 10th out of 23 detectors (see Table \ref{tab:combined_results}). This model effectively combined fine-tuned RoBERTa Base and RoBERTa Base OpenAI models, with perplexity-based weighting to give more influence to lower-perplexity models, enhancing overall accuracy. It consistently delivered strong results across various generator models, including ChatGPT and GPT-3.

In the adversarial testing, the fine-tuned RoBERTa model outperformed other detectors, achieving an aggregate (AGG) score of 0.801 and ranking 8th out of 22 (see Table \ref{tab:combined_results}). This demonstrates the model's robust adaptability in adversarial conditions, achieving top scores with GPT-3 and GPT-4, even under altered input scenarios.

\subsection{Model Comparison}
The performance of various detectors was evaluated under both non-adversarial and adversarial conditions, revealing key insights into their strengths and limitations.

In the non-adversarial setting, FT RoBERTa + RoBERTa OpenAI emerged as the top performer, achieving an AGG TPR of 0.826 and ranking 10th overall. It demonstrated exceptional performance with models such as ChatGPT (TPR: 0.960) and GPT-4 (TPR: 0.861), outperforming FT RoBERTa (AGG TPR: 0.813, ranked 12th) and the ensemble model FT RoBERTa + RoBERTa OpenAI + BERT (AGG TPR: 0.825, ranked 11th). Interestingly, Binoculars \cite{hans2024spottingllmsbinocularszeroshot} showed strong results with specific generators like GPT-3 (TPR: 0.989) and Llama-Chat (TPR: 0.973). However, its inconsistent performance with other generators, such as MPT (TPR: 0.447), limited its reliability. In contrast, FT RoBERTa + RoBERTa OpenAI demonstrated stable results across all generators, including strong performances with Mistral-Chat (TPR: 0.936) and Cohere-Chat (TPR: 0.748), underscoring its robustness and versatility.

Under adversarial conditions, FT RoBERTa proved to be the most robust model, achieving an AGG TPR of 0.801 and ranking 8th overall. It excelled with GPT-3 (TPR: 0.873) and MPT-Chat (TPR: 0.861), outperforming FT RoBERTa + RoBERTa OpenAI (AGG TPR: 0.760, ranked 10th) and FT RoBERTa + RoBERTa OpenAI + BERT (AGG TPR: 0.749, ranked 11th). In comparison, the SuperAnnotate AI Detector \cite{superannotate2024} delivered competitive results with ChatGPT (TPR: 0.963), but its performance was inconsistent, particularly with GPT-2 (TPR: 0.411) and Mistral (TPR: 0.342). These results emphasize the variability of some detectors when faced with adversarial data, highlighting the consistent reliability of FT RoBERTa.

The consistent dominance of FT RoBERTa + RoBERTa OpenAI in non-adversarial settings and FT RoBERTa in adversarial conditions underscores the importance of tailoring architectures to specific scenarios. While models like Binoculars and SuperAnnotate excelled in isolated cases, their lack of stability across diverse generators reinforces the value of robust, well-balanced models like FT RoBERTa. These findings suggest that future efforts should focus on further optimizing architectures to enhance cross-domain robustness and adversarial detection capabilities.

\section{Limitations}

Our approach, while effective, has several limitations. Focusing on RoBERTa models for fine-tuning and ensemble weighting excluded alternatives like RemBERT \cite{DBLP:conf/iclr/ChungFTJR21} and XLM-RoBERTa \cite{DBLP:journals/corr/abs-1911-02116}, which might better handle longer sequences, noisy data, and multi-label tasks.  

Due to computational constraints, we trained on a subset of the RAID dataset, limiting the model's ability to capture its full diversity. Training on the full dataset could greatly improve detection performance, especially for underrepresented domains.

Performance variability across generator models (e.g., GPT-4 vs. Mistral) and limited multilingual capabilities highlight the need for better cross-domain generalization and robust multilingual detection. While the ensemble approach enhanced generalization, it increased computational overhead, warranting exploration of more efficient strategies in future work.

\section{Discussion and Conclusion}

In this paper, we proposed an ensemble-based approach for cross-domain MGT detection, combining fine-tuned RoBERTa Base and RoBERTa Base OpenAI detectors with inverse perplexity weighting. Our method achieved competitive results, ranking 10th and 8th in non-adversarial and adversarial tasks, respectively, in Task 3 of the GenAI content detection workshop at COLING-2025. Inverse perplexity weighting improved generalization by prioritizing more confident models across diverse domains. For non-adversarial tasks, we explored an inverse perplexity-based ensemble approach. However, the detectors in this ensemble underperformed compared to the fine-tuned RoBERTa model, highlighting the value of fine-tuning on task-specific data and suggesting avenues for refining ensemble techniques.

Our results show that transformer-based models, particularly RoBERTa, are effective for non-adversarial and adversarial MGT detection. For non-adversarial detection (Subtask A), we achieved a score of 0.826, and for adversarial detection (Subtask B), we scored 0.801. However, cross-domain detection remains challenging, especially with varied generator models and multilingual data. Our system performed well with generators like ChatGPT and GPT-4 but struggled with others like Cohere and Mistral, indicating the difficulty of detecting diverse machine-generated content.

Due to limited computational resources, we trained on a subset of the available data. Despite this, our models performed well, demonstrating the potential of our approach even with partial data. This work lays the foundation for further progress in MGT detection, especially in adversarial and cross-lingual settings. Future research can focus on enhancing multilingual capabilities, incorporating more diverse language models, and exploring dynamic ensemble strategies to improve performance across domains and attack scenarios.

\bibliography{coling_latex}

\appendix

\section{Appendix}
\label{sec:appendix}

\begin{table}[h]
    \centering
    \small
    \renewcommand{\arraystretch}{1.25} % Adjust the row spacing
    \setlength{\tabcolsep}{2.5pt} % Adjust column separation for clarity
    \begin{tabular}{p{4cm}|>{\centering\arraybackslash}p{3cm}}
        \hline \hline
        \textbf{Tools \& Libraries} & \textbf{Version} \\
        \hline
        Python & 3.10.14 \\
        Pandas & 2.2.2 \\
        NumPy & 1.26.4 \\
        PyTorch & 2.4.0 \\
        Transformers & 4.44.2 \\
        Evaluate & 0.4.3 \\
        WandB & 0.16.6 \\
        \hline
    \end{tabular}
    \caption{Main tools and libraries used in our system}
    \label{tab:library_versions}
\end{table}

Table \ref{tab:library_versions} provide the details about the
corresponding libraries, which are beneficial to help replicate
our experiments.

\end{document}